%% file: main.tex
\documentclass[10pt,twocolumn,letterpaper]{article}

\usepackage[pagenumbers]{cvpr} 
\usepackage{graphicx}
\usepackage{booktabs}
\usepackage{multirow}
\usepackage{array}
\usepackage{xcolor}

\definecolor{cvprblue}{rgb}{0.21,0.49,0.74}
\usepackage[pagebackref,breaklinks,colorlinks,allcolors=cvprblue]{hyperref}

\def\paperID{14780} 
\def\confName{CVPR}
\def\confYear{2025}

\title{Dynamic-VLM: Simple Dynamic Visual Token Compression for VideoLLM}

\author{
    Han Wang \hspace{1em} 
    Yuxiang Nie \hspace{1em} 
    Yongjie Ye \hspace{1em} 
    Deng GuanYu \hspace{1em} 
    Yanjie Wang \hspace{1em} \\
    Shuai Li \hspace{1em} 
    Haiyang Yu \hspace{1em} 
    Jinghui Lu \hspace{1em} 
    Can Huang \\
    \vspace{0.5em} \\ 
    Bytedance Inc. 
}

\newcommand{\model}[1]{Dynamic-VLM}

\begin{document}
\maketitle
\input{sec/0_abstract}    
\input{sec/1_intro}
\input{sec/2_related}
\input{sec/3_methods}
\input{sec/4_experiments}
\input{sec/5_conclusions}
{
    \small
    \bibliographystyle{ieeenat_fullname}
    \bibliography{main}
}


\end{document}

%% file: sec/0_abstract.tex
\begin{abstract}
The application of Large Vision-Language Models (LVLMs) for analyzing images and videos is an exciting and rapidly evolving field. In recent years, we've seen significant growth in high-quality image-text datasets for fine-tuning image understanding, but there is still a lack of comparable datasets for videos. Additionally, many VideoLLMs are extensions of single-image VLMs, which may not efficiently handle the complexities of longer videos.
In this study, we introduce a large-scale synthetic dataset created from proprietary models, using carefully designed prompts to tackle a wide range of questions. We also explore a dynamic visual token compression architecture that strikes a balance between computational efficiency and performance. Our proposed \model{} achieves state-of-the-art results across various video tasks and shows impressive generalization, setting new baselines in multi-image understanding. Notably, \model{} delivers an absolute improvement of 2.7\% over LLaVA-OneVision on VideoMME and 10.7\% on MuirBench.
\end{abstract}

%% file: sec/1_intro.tex
\section{Introduction}
\label{sec:intro}
\begin{figure}
    \centering
    \includegraphics[width=\linewidth]{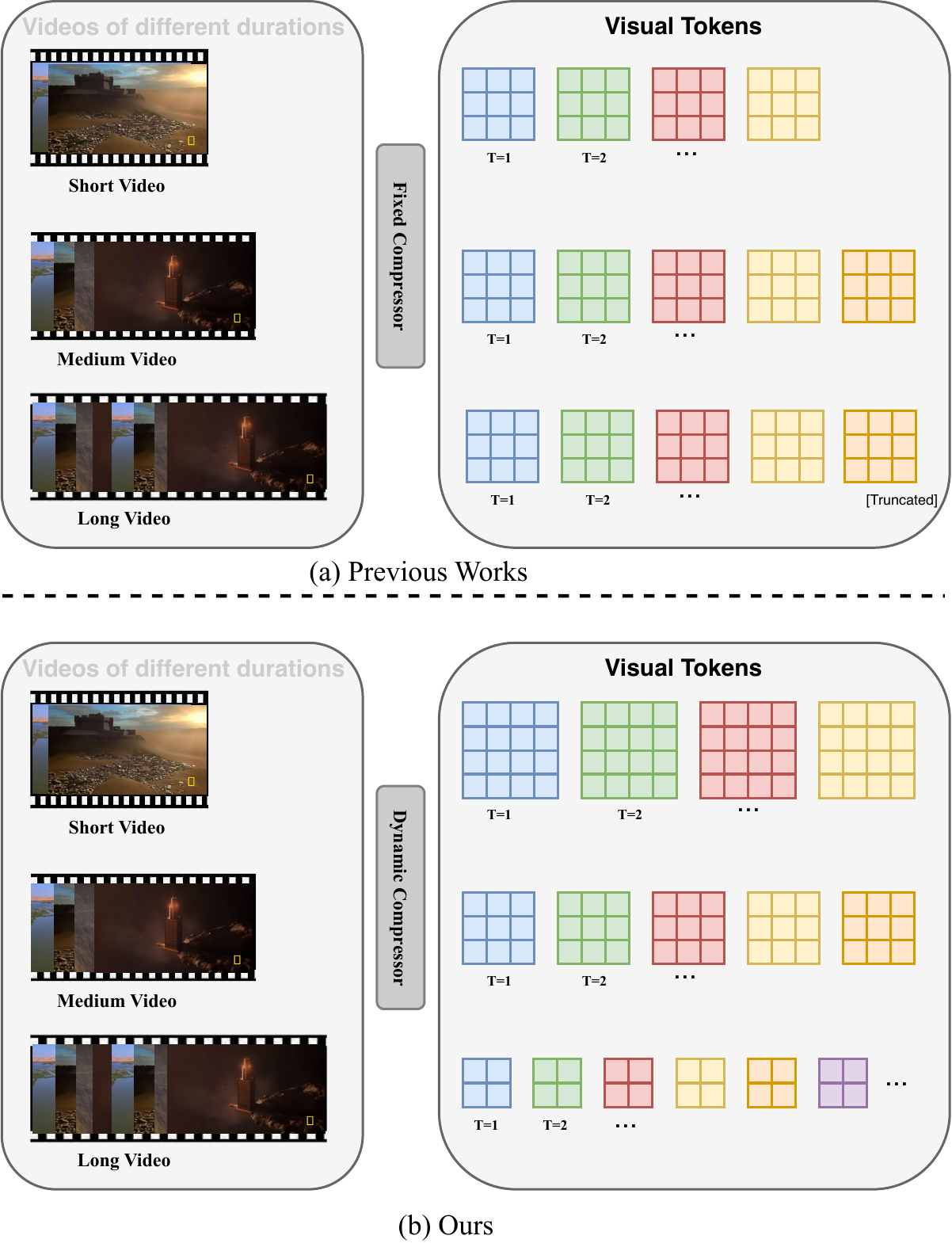}
    \caption{Demonstration of previous works and our \model{}. We use a flexible token compressor for visual content, enabling us to show videos of different lengths with varying token counts. For short videos, we keep tokens uncompressed to provide detailed information, and for long videos, we use a high compression ratio to enhance temporal details. For the sake of simplicity, visual encoders are excluded from the illustration.}
    \label{fig1}
\end{figure}
In recent years, there has been significant success in the field of Large Vision and Language Models (LVLM)~\cite{llava,minigpt,minigpt2,videollava,elysium}. Leveraging the impressive capabilities of Large Language Model (LLM)~\cite{vicuna2023,llama}, along with a well-trained visual encoder~\cite{siglip, clip} and a large data scale, enables these models to comprehend both textual and visual information. The extension of instruction following ability from LLM to LVLM is also a notable development.
The state-of-the-art close-sourced LVLM, such as GPT-4V~\cite{gpt4v}, GPT-4o~\cite{gpt4o}, Gemini-1.5-Pro~\cite{gemini15} usually have the capacities to process both vision and language tasks. Driven by large-scale multi-modal data, these models acquire general abilities across different tasks, such as image QA and video QA. Additionally, benefiting from long-context processing, they can handle extremely long videos, encompassing thousands of frames. 

Similarly, within the open community, in the field of single images, it has proven effective in gathering large-scale data from closed-source models. These datasets~\cite{llava,llava15,allava,sharegpt4v,sharegpt4o} are commonly utilized by most state-of-the-art methods.
In the realm of Video Large Language Models (VideoLLMs), where large-scale synthetic data is lacking, many studies have focused on leveraging open-resource data typically derived from previous datasets to enhance the capabilities of LVLMs in processing videos of various lengths. However, due to the short context of LLMs and the low quality of video data, a significant gap still exists between industry-level models and closed-source alternatives. 
Several efforts have been made for short and long video understanding. Some works~\cite{moviechat,slowfastllava} explored storing temporal information in an external memory module. These models can be seen as merging the features of both different frames and different tokens into one place. However, these methods often struggle significantly due to the loss of detailed information, leading to inferior performance compared to other approaches. This is particularly noticeable when dealing with tasks that demand frame-level details, such as OCR capabilities. In this year, a trend is observed that involves simply scaling the context of VideoLLM by tuning with long video cases. As shown in Fig.~\ref{fig1}, these methods~\cite{longva,longllava} apply the same compressor for long videos and shorter ones, keeping the compression ratio fixed. However, these methods suffer quality degradation when employing an extensive context. As a result, the cutting-edge long VideoLLM only achieves comparable or even inferior results compared to standard settings. This raises a question: How can the context window and processed frames be balanced effectively?

To address the aforementioned issues, this study focuses on two main aspects: 1) Collecting question-answering pairs from closed-source models, specifically GPT-4V~\cite{gpt4v} and GPT-4o~\cite{gpt4o}. By carefully selecting raw videos and crafting prompts, we leverage the capabilities of these proprietary models to gather a wide range of video question-answering pairs. 2) Regarding the architecture of VideoLLM, we investigate the implementation of a more dynamic representation for a single image. Instead of using a fixed number of tokens, we explore utilizing a variable number of tokens to represent each image. This approach allows for flexibility in handling videos of varying lengths during inference. We evaluate our proposed \model{} across a diverse set of tasks, including Open-ended VideoQA, Multiple-choice VideoQA, and a broader Multi-image QA that lies outside the training data domain of \model{}. The top performances on all benchmarks demonstrate the effectiveness of \model{} and the proposed dataset.

Our contributions can be summarized in three key aspects:
\begin{itemize}
\item We create a synthetic high-quality video-text dataset for training VideoLLMs.
\item We introduce a versatile visual token compressor designed to accommodate videos of different lengths. We explore how adjusting the number of visual tokens per image and visual context can impact the ultimate outcomes.
\item We conduct extensive experiments on various tasks to prove the effectiveness of \model{}.
\end{itemize}

%% file: sec/2_related.tex
\section{Related Works}
\label{sec:relatedworks}
\subsection{Architecture Design of VideoLLM}
\label{videollm}
In the early stages of VideoLLM development, many approaches could only handle a limited number of frames, restricting their usability to short videos. Following the design principles of single-image LVLMs, these methods often employed fixed compressors or simple pooling operations along the temporal axis for videos of varying lengths.
To address the challenge of processing longer videos, certain studies~\cite{moviechat,slowfastllava} introduced an external memory module to retain temporal information. However, this approach may overlook detailed frame-level information such as textual content. In contrast, other works~\cite{longva,longllava} chose to scale the context without considering the increasing computational cost and potential performance degradation that comes with expanding the context window of LLM without adequate training.
Therefore, it is natural to explore a more flexible approach that balances context and the processing of additional frames, leading to a more adaptable token representation for each image in a video.
\subsection{Synthetic Data Construction}
In the LVLM domain, it is a common practice to incorporate synthetic data generated by large closed-source models. For instance, LLaVA \cite{llava} generates a substantial volume of visual instruction tuning data by leveraging GPT-4. This involves transforming the original captions and additional information into prompts for GPT-4~\cite{gpt4roi} to generate question-answering pairs closely related to the visual content. ShareGPT4V~\cite{sharegpt4v} and ShareGPT4o~\cite{sharegpt4o} directly extract detailed caption data from GPT-4V~\cite{gpt4v} and GPT-4o~\cite{gpt4o}, respectively, enhancing the performance of LVLM within the open community. ALLaVA~\cite{allava} not only aggregates detailed captions but also QA pairs from GPT-4V~\cite{gpt4v}, amassing a total of 1.4 million data points that have found wide application in subsequent research endeavors.

However, in the realm of video, the fine-tuning of current VideoLLMs still heavily relies on low-quality data sourced from existing datasets~\cite{perceptiontest,yu2019activityqa,nextqa}, which often encompass only a limited range of questions like activity recognition or object counting. This leads to a significant gap in the quality and diversity of available data, thereby impacting the effectiveness of model training and performance in video-related tasks.

%% file: sec/3_methods.tex
\section{Model Architecture}
\begin{figure*}
    \centering
    \includegraphics[width=\linewidth]{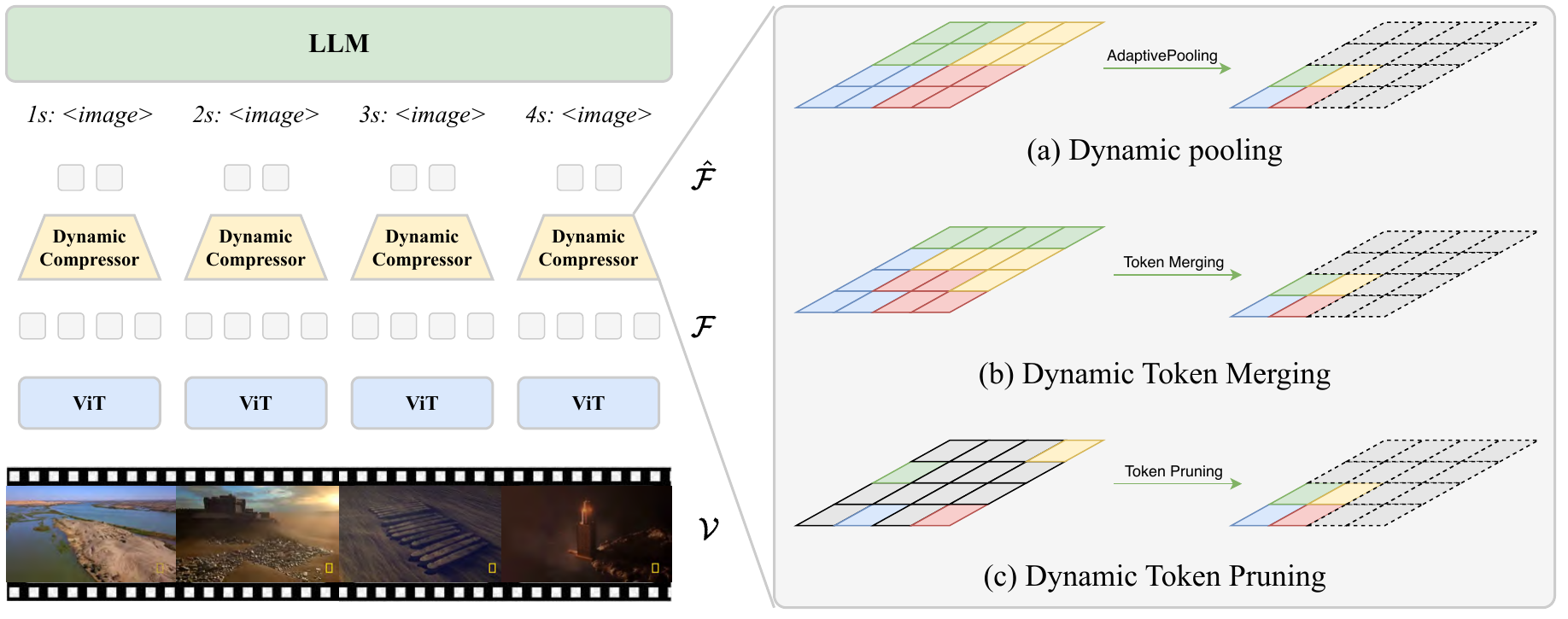}
    \caption{For each video, we independently extract visual tokens for each key frame using a ViT. These visual tokens are then compressed using dynamic compressors before being input to the LLM, along with timestamp text and instructions. We discuss three potential candidates for dynamic compressors.}
    \label{fig:framework}
\end{figure*}

\subsection{Overall Architecture}
As shown in Figure~\ref{fig:framework}, following the most common architecture in the open-sourced LVLM, \model{} consists of three main parts: a vision encoder, an LLM, and a connector to bridge the modalities. Typically, we use CLIP-ViT-Large@336p~\cite{clip} as the default vision encoder and Qwen-2.5 series~\cite{qwen25} as the default LLMs. 
\subsection{Dynamic Visual Token Compressors}
As discussed in Sec.~\ref{videollm}, in this work, we would like to employ a more flexible token compressor for VideoLLM to adapt to different lengths of videos.
%
In video scenarios, the processing of a significant number of frames is often necessary, creating a challenge in striking a balance between visual tokens per image and the input frame count. Previous studies have attempted to address this issue by expanding the context of LVLM, enabling the accommodation of more frames. However, these methods typically incur substantial computational costs and result in diminished performance. Therefore, it is prudent to explore more adaptable strategies for visual token encoding, allowing for the utilization of additional tokens to represent images in brief videos while reducing token usage for longer videos. In essence, the approach involves employing an increased number of tokens to depict a single image in short videos and navigating a trade-off between accommodating more frames and enhancing the tokens per frame.
\\
Given an input video $\mathcal{V}=\{\mathbf{X}_0,...,\mathbf{X}_{N-1}\}$ with $N$ frames, each frame $\mathbf{X}_i$ is independently input into the vision encoder to extract corresponding visual features $\mathbf{F}_i$. Thus, the output of the vision encoder can be represented as $\mathcal{F}=\{\mathbf{F}_0,...,\mathbf{F}_{N-1}\}$, where $\mathbf{F}_i\in \mathbb{R}^{(H\times W) \times C}$ denotes the visual tokens of the $i$-th frame. Then we apply a dynamic token compressor to reduce the token count: $\hat{\mathcal{F}}=\texttt{Compressor}(\mathcal{F}, M)$, where $\hat{\mathcal{F}}=\{\hat{\mathbf{F}}_0,...,\hat{\mathbf{F}}_{M-1}\}$, $M$ is the number of tokens that we want to retain.
We discuss three simple potential dynamic token processors:

1) Dynamic Spatial Pooling. It is natural to apply a pooling operation as a down-sampling method, which can be simply written as:
\begin{equation}
\hat{\mathcal{F}}=\texttt{AdaptiveAvgPool2d}(\mathcal{F}, [H, W]), 
\end{equation}
where $H=W \in [4, 28]$, so the final number of tokens $M=H\times W$.

2) Dynamic Token Merging. ToMe~\cite{tome} stands for the merging operation in different tokens based on the cosine similarity score in ViT, which is also applied by MovieChat~\cite{moviechat}. Here, we simply apply the k-th Bipartite Soft Matching to get the similar tokens:
\begin{equation}
\hat{\mathcal{F}}=\texttt{k-th-Bipartite-Soft-Matching}(\mathcal{F}, k), 
\end{equation}
where $k=\frac{N}{M} \in [1, 49]$.

3) Pruning is a type of vision token compression method introduced in a previous study~\cite{elysium} for object-level perception tasks using LVLM, allowing for the use of any number of tokens to represent one image. This can be expressed as:
\begin{equation}
\hat{\mathcal{F}}=\texttt{KeepTopK}(\texttt{GumbelSoftmax}(\texttt{MLP}(\mathcal{F}), M, \mathcal{F})), 
\end{equation}
where $M \in [16, 576]$.
\section{Video QA Data Construction}
\begin{figure}
    \centering
    \includegraphics[width=\linewidth]{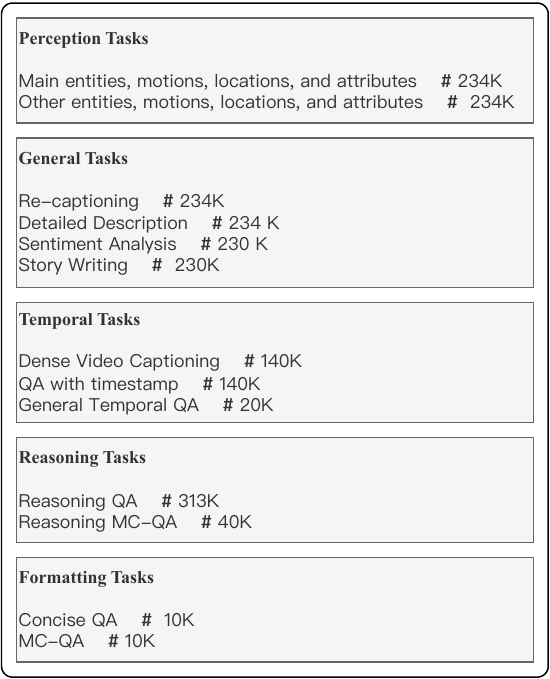}
    \caption{Distribution of different tasks in our synthetical data.}
    \label{fig:data}
\end{figure}
\begin{table*}[ht]
\centering
\begin{tabular}{l|c|c|c|c|c}
\toprule
\textbf{Datasets} & \textbf{\# Samples} & \textbf{Domain} & \textbf{Video Duration} & \textbf{Tasks} & \textbf{Annotated by} \\ 
\midrule
LLaVA~\cite{llava} & 158K & Image & N.A. & Mixed & GPT4\\
Q-Instruct~\cite{qinstruct} & 200K & Image & N.A. & General VQA & GPT3.5\\
ShareGPT4V~\cite{sharegpt4v} & 100K & Image & N.A. & Detailed Caption & GPT4-V\\
ShareGPT-4o	\cite{sharegpt4v} & 220K & Image \& Video \& Audio & Short & Detailed Caption & GPT4-o\\
ALLaVA~\cite{allava} & 1.4M & Image & N.A. & Mixed & GPT4-V\\
\midrule
VideoChatGPT~\cite{videochatgpt} & 100K & Video & Short & General VQA & Human \& GPT3.5\\
VideoGPT+~\cite{videogptp} & 112K & Video & Short & General VQA & GPT4-o \& GPT3.5\\
ShareGPT4Video~\cite{sharegpt4video} & 40K & Video & Short & Detailed Caption & GPT4-o\\
\midrule
\model{} (\textit{ours}) & 2M & Video & Mixed & Mixed & GPT4-V \& GPT4-o\\
\bottomrule
\end{tabular}
\caption{Comparison of different datasets. Short video duration means the datasets does not contain hour-length videos.}
\label{datasets_compare}
\end{table*}
\subsection{Raw Data Preparation}
We aim to create a more general dataset rather than merely enhancing model performance on specific benchmarks. Different from previous approaches~\cite{videogptp,videochat2}, we avoid using overlapping videos from existing benchmarks or their training splits. Instead, we utilize raw videos sourced from three existing datasets: WebVid-10M~\cite{webvid}, InternVid-10M~\cite{internvid}, and HDVILA-100M~\cite{hdvila}, all of which have not been used in any benchmarks. For the WebVid-10M and InterVid-10M datasets, a majority of the videos are short in length, and there are instances of similar captions appearing across different videos. To address this, we initially remove video duplicates by utilizing only the original captions. Following this, we implement a data filtration pipeline similar to the one proposed in LLaVA~\cite{llava}. This involves parsing each caption into raw noun chunks, downsampling high-frequency chunks to reduce the number of videos. As a result, we obtain approximately 547k videos from InterVid-10M and 349k videos from WebVid-10M.
When prompting the closed-source model to generate question-answering pairs, for the 349k videos in WebVid-10M, we provide both original captions and videos. However, for the 547k videos in InterVid-10M, due to the lower quality of the generated captions, only the videos are provided. Sampling for these datasets is conducted at a rate of 1 frame per second (1FPS).
For HDVILA-100M, we use the original 3.3M videos instead of using the clipped 100M videos for including more longer videos. We did not process the raw videos because we find that the original videos have enough diversity. In most cases, we sample up to 128 frames per video if the duration exceeds 128 seconds; otherwise, we sample at 1 FPS. Additionally, a few videos are sampled up to 256 frames to enhance the model's ability to process longer video sequences.
\subsection{Task Design}
Designing prompts that address a wide range of tasks while minimizing costs is essential. To achieve this, we utilize different prompts tailored to specific datasets. For WebVid-349k and InternVid-547k, which feature shorter video durations and lower resolutions, the API call costs are relatively low. As a result, we create prompts that cover a broad spectrum of diverse questions. In contrast, for videos from HDVILA, which typically have longer durations and richer visual content, we design prompts that encourage in-depth reasoning for visual question answering, closely tied to the video's visual content. As shown in Figure~\ref{fig:data}, we categorize our prompts into several categories:
\begin{itemize}
\item \textbf{Perception Tasks:} Here, we define perception tasks to identify object entities in videos along with their attributes, locations, and motions. In our initial investigations, we observed that the closed-source model tends to output vague terms like ``a person" or ``an item," likely due to safety preferences. To address this, we instruct the model to avoid using such ambiguous language. Furthermore, we allow the model to refrain from providing an answer when there is insufficient information.
\item \textbf{General Tasks:} This category encompasses re-captioning tasks, detailed descriptions, sentiment analysis, and story writing, all aimed at enhancing the model's performance on general tasks, with a particular emphasis on language-related tasks.
\item \textbf{Temporal Tasks:} Videos have an additional temporal dimension compared to images, which requires the design of QA pairs that emphasize temporal awareness. We formulate three types of questions: dense video captioning, QA with timestamp, and general temporal QA. In the first two types, we incorporate explicit timestamps, such as \texttt{<Frame 3>}, in the question, the answer, or both. In the third type, we refrain from using specific timestamps and instead pose more general queries about temporal events.
\item \textbf{Reasoning Tasks:} Reasoning tasks are a crucial component of our training data, significantly improving the model's final performance. The reasoning process often involves intricate details within the visual modality, further enhancing the general capabilities of \model{}.
\item \textbf{Formatting Tasks:} We also collect a smaller set of formatted data to guide the model's output in accordance with the requirements of multi-choice QA and to provide concise answers for open-ended QA. This facilitates the evaluation of \model{}'s performance, particularly on benchmarks that utilize multiple-choice QA. We apply the prompt ``\{question\}Answer with only one letter” to encourage \model{} to generate responses in the correct format, and we find this approach to be very effective.
\end{itemize}
Next, we filter out corrupted responses and answers that do not conform to the specified format or are rejected by the closed-source models for various reasons. We convert the timestamps in the QA pairs into a more natural format, such as changing \texttt{<Frame 3>} to "frame of 3s." We save the sampled frames directly, rather than the original video, to ensure that our \model{} views the exact same frames as the closed-source model.
In the end, we gather about 2 million question-answering pairs.
\subsection{Datasets Comparison}
As shown in Table \ref{datasets_compare}, compared to existing video datasets, our proposed dataset features a significantly larger scale, variable durations, and a diverse range of tasks. Leveraging raw data from HDVILA, we work with exceptionally lengthy videos that span hours. For these extended videos, we extract up to 256 frames to ensure robustness. These sampled frames in long video often capture distinct scenes, which can be advantageous for VideoLLM in capturing intricate details within each frame.
\section{Training}
\begin{table*}[ht]
    \centering
    \setlength{\tabcolsep}{5pt}
    \begin{tabular}{l|c|c|cc|cc|cc|cc}
        \toprule
        \multirow{2}{*}{\textbf{Model}} & \multirow{2}{*}{\textbf{LLM Size}} & \multirow{2}{*}{\textbf{API Ver.}} & \multicolumn{2}{c|}{\textbf{MSVD-QA}} & \multicolumn{2}{c|}{\textbf{MSRVTT-QA}} & \multicolumn{2}{c|}{\textbf{TGIF-QA}} & \multicolumn{2}{c}{\textbf{ActivityNet-QA}}\\
        \cline{4-11}
        &      &             & Acc.  & Score     & Acc.  & Score     & Acc.  & Score     &   Acc.    & Score \\
        \midrule
        Vista-LLaMA \cite{vistallama} & 7B & N/A & 65.3 & 3.6 & 60.5 & 3.3& -& - & 48.3&  3.3\\
        Video-LLaMA2 \cite{cheng2024videollama} & 7B & N/A & 70.9 & 3.8 & - & - & - & - & 50.2 & 3.3\\
        MovieChat~\cite{moviechat} &7B & N/A & 75.2      & 3.8       & 52.7      & 2.6       & -      & -     &               45.7       & 3.4\\
        \midrule
        VideoLLaMA~\cite{videollama} & 7B & 0301 & 51.6      & 2.5       & 29.6      & 1.8       & -      & -     &               12.4       & 1.1 \\
        LLaMA-Adapter~\cite{llamaadapter} &7B & 0301  & 54.9      & 3.1       & 43.8      & 2.7       & -      & -     &               34.2       & 2.7 \\
        VideoChat~\cite{videochat} &7B & 0301 & 56.3      & 2.8       & 45.0      & 2.5       & 34.4      & 2.3     &               26.5       & 2.2 \\
        VideoChatGPT~\cite{videochatgpt}&  7B     & 0301 & 64.9      & 3.3       & 49.3      & 2.8       & 51.4      & 3.0     &               35.2       & 2.7 \\
        Valley-v3~\cite{valley} &7B & 0301 & 60.5      & 3.3       & 51.1      & 2.9       & -      & -     &               45.1       & 3.2 \\
        Elysium~\cite{elysium} & 7B & 0301 & 75.8 & 3.7 & 67.5 & 3.2 & 66.6 & 3.6&  43.4 & 2.9 \\
        IG-VLM~\cite{igvlm} & 7B & 0301 &  78.8& 4.1 & 63.7& 3.5 & 73.0& 4.0 & 54.3& 3.4\\
        FreeVA \cite{freeva} & 7B & 0301 & 81.5 & 4.0 & 72.9 & 3.5 &-& -&  58.3 & 3.5\\
        \midrule
        Chat-UniVi \cite{chatunivi} & 7B & 0613 & 65.0 & 3.6 & 54.6 & 3.1 & - & - & 45.8 & 3.2\\
        BT-Adapter \cite{btadapter} & 7B & 0613 & 67.5 & 3.7 & 57.0 & 3.2 & -& -& 45.7 & 3.2\\
        LLaMA-VID \cite{llamavid} & 7B & 0613 & 69.7 & 3.7 & 57.7 & 3.2 &-& -&  47.4 & 3.3\\
        LLaVA-NeXT-Interleave  \cite{llavanextinterleave} &  7B & 0613 &  -& - & -& - & -& - & 55.3 & 3.1\\
        LLaVA-NeXT-Video \cite{zhang2024llavanextvideo} &  7B & 0613 &  -& - & -& - & -& - & 53.5 & 3.2\\
        LLaVA-NeXT-Video \cite{zhang2024llavanextvideo} &  7B & 0613 &  -& - & -& - & -& - & 60.2 & 3.5\\
        ST-LLM \cite{stllm} & 7B & 0613 & 74.6 & 3.9 & 63.2 & 3.4 & - & - &50.9 & 3.3\\
        FreeVA \cite{freeva} & 7B & 0613 & 71.2 & 3.8 & 58.5 & 3.2 & - & - & 53.7 & 3.5\\
        VideoGPT+ \cite{videogptp} &  7B & 0613 &72.4 & 3.9 & 60.6 & 3.6 & 74.6 & 4.1 & 50.6 & 3.6\\
        \midrule
        \textbf{\model{}} (\textit{ours}) &  7B & 0613 & 76.0 & 3.8 & 64.5 & 3.7 & 76.6 & 3.8 & 63.9 & 3.4\\
        \bottomrule
    \end{tabular}
    \caption{Results on open-ended VideoQA. Many studies do not explicitly state the GPT version used, so we conducted searches within their respective GitHub repositories to obtain this information. For methods where we could not find the version in either the paper or the repository, we indicate the version as ``N/A." It is important to note that the API version can significantly impact the final metrics (by over 10 absolute points, according to FreeVA). Therefore, it is preferable to compare models using the same API version for a fair evaluation.}
    \label{tab:openvideoqa}
\end{table*}
\begin{table*}[ht]
\centering
\begin{tabular}{l|c|c|c|c|c|c}
\toprule
\textbf{Methods} & \textbf{LLM Size} & \textbf{VideoMME} & \textbf{MLVU} & \textbf{TempCompass} & \textbf{EgoSchema} & \textbf{PerceptionTest}\\ 
\midrule
GPT-4V~\citep{gpt4v}& N/A & 59.9\%/63.3\%& 49.2\% & - & - &- \\
GPT-4o~\citep{gpt4o}& N/A & 71.9\%/77.2\%& 64.6\% & 71.0\% & - & - \\
Gemini-1.5-Pro~\citep{gemini15} & N/A & 75.0\%/81.3\%& - & - & - & - \\
GPT-4o mini~\citep{gpt4omini} & N/A & 64.8\%/68.9\%& - & - & - & -  \\
\midrule
LLaVA-Next-Video~\citep{zhang2024llavanextvideo} & 32B & 60.2\%/63.0\%& 65.5\%& 68.7\% & 60.9\% & 59.4\% \\ 

VILA-1.5~\citep{lin2023vila} & 40B & 60.1\%/61.1\% & 56.7\% & - & 58.0\% & 54.0\% \\
VITA~\citep{vita} & 8$\times$7B & 55.8\%/59.2\% & - & - & - & -  \\
VideoLLaMA2~\citep{cheng2024videollama} & 72B & 61.4\%/63.1\% & - & - & 63.9\% & 57.5\%\\
LLaVA-OneVision~\citep{li2024llavaov} & 72B & 66.2\%/69.5\% & 68.0\% & - & 62.0\% & 66.9\%\\
\midrule
VideoChat2-Mistral~\citep{li2024mvbench} & 7B & 39.5\%/43.8\% & - & - & 54.4\% & - \\
Video-LLaMA2~\citep{cheng2024videollama} & 7B & 47.9\%/50.3\% & 48.5\% & - & - & 51.4\%\\
LongVA~\citep{zhang2024long}& 7B &  52.6\%/54.3\% & 56.3\% & 56.1\% & - & -\\
Long-LLaVA~\citep{longllava} & 7B &  52.9\%/57.1\% & - & - & - & - \\
IXC-2.5~\citep{zhang2024internlm}& 7B &  55.8\%/58.8\% & 37.3\% & 61.3\% & - & 34.4\%\\
Kangaroo~\citep{liu2024kangaroo} & 8B & 56.0\%/57.6\% & 61.0\% & - & - & -\\
VideoCCAM~\citep{fei2024videoccam} & 9B & 53.9\%/56.1\% & - & - & - & - \\
VideoCCAM~\citep{fei2024videoccam} & 14B & 53.2\%/57.4\% & - & - & - & - \\
LLaVA-OneVision~\citep{li2024llavaov} & 7B & 58.2\%/61.5\% & 64.7\% & 64.8\% & 60.1\% & 57.1\% \\
\midrule
\model{} (\textit{ours}) & 7B & 60.9\%/64.0\% & 65.0\% & 62.2\% & 68.6\% & 68.8\% \\
\model{} (\textit{ours}) & 14B & 64.6\%/68.8\% & 70.1\% & 66.2\% & 75.2\% & 72.1\%\\
\bottomrule
\end{tabular}
\caption{Results on Multi-choice VideoQA. We compare our \model{} with closed-source models, larger-scale models, and comparable-scale models. Our \model{} achieves the best performance among models ranging from 7B to 14B parameters, even surpassing some larger-scale models and demonstrating comparable performance to GPT-4o mini on VideoMME (w.o./w. subtitles). ``N/A" indicates that the model is closed-source.}
\label{tab:mcvideoqa}
\end{table*}
\begin{table*}[ht]
    \centering
    \setlength{\tabcolsep}{12pt}
    \begin{tabular}{l|c|c|c|c}
        \toprule
        \textbf{Model} & \textbf{LLM Size} & \textbf{Training Domain} & \textbf{MuirBench} & \textbf{MMIU} \\
        \midrule
        Human & N/A & N/A & 93.2\% & -\\
        \midrule
        GPT-4o \cite{gpt4o} & N/A & N/A & 68.0\% & 55.7\% \\
        GPT-4V \cite{gpt4v} & N/A & N/A & 62.3\% & - \\
        \midrule
        ShareGPT4V \cite{sharegpt4v} & 7B & Image & - & 18.5\%\\
        LLaVA-v1.5 \cite{llava15} & 7B & Image & 23.5\% & 19.2\%\\
        LLaVA-NeXT-Interleave \cite{llavanextinterleave} & 7B & Image \& Multi Image & - & 32.4\%\\
        LLaVA-OneVision \cite{li2024llavaov} & 7B & Image \& Multi Image \& Video & 41.8\% & -\\
        LLaVA-v1.5 \cite{llava15} & 13B & Image & 24.4\% & -\\
        \midrule
        \model{} (\textit{ours}) & 7B & Image \& Video & 50.7\% & 43.7\%\\
        \model{} (\textit{ours}) & 14B & Image \& Video & 52.5\% & 51.7\%\\
        \bottomrule
    \end{tabular}
    \caption{Results on Multi-choice Multi-image QA. All benchmarks are completely out of the domain of \model{}’s training data. Nevertheless, \model{} demonstrates strong zero-shot performance on these tasks, even surpassing models that have been trained on multi-image data.}
    \label{mcmiqa}
\end{table*}
\subsection{Pretraining Stage}
Following previous works~\cite{llava15,llavanext,li2024llavaov}, we pretrain \model{} on llava-558K, initially only unfreezing the compressors to establish good initial parameters. We then train \model{} end-to-end on a mixed dataset for detailed captioning, which includes high-quality sources such as ALLAVA~\cite{allava}, ShareGPT4V~\cite{sharegpt4v}, and LLaVA-Recap-CC3M~\cite{llavanext}. We filter out annotations that contain repetitive patterns in the data. In this phase, we unfreeze all parameters in \model{} to enhance its overall visual perception capabilities. In all stages, we use a simple system prompt: ``You are a helpful visual assistant.", a learning rate of $2\times 10^{-5}$  is applied with a batch size of 128 for both the LLM and compressors, while a learning rate of $\frac{1}{5}$
  of this rate is used for the ViT. In training stages with images, the number of tokens per frame is randomly sampled from the interval $ \left[ 16, 576\right]$. 
\subsection{Visual Instruction Tuning}
The image data proves to be very effective contributing to the VideoLLM~\cite{li2024llavaov,llavanext}. Thus we involve a large scale of public-accessible data to enhance the visual instruction ability of \model{}, Specifically, we gather datasets from mainly two categorizes.
\begin{itemize}
    \item General VQA: ALLaVA \cite{allava}, Q-Instruct \cite{qinstruct}, VQAv2 \cite{vqav2}, GQA \cite{gqa}, A-OKVQA \cite{aokvqa}, ShareGPT4o \cite{sharegpt4o}, VizWiz \cite{vizwiz}, Visual7W \cite{visual7w}, IDK \cite{idk}, OKVQA \cite{okvqa}, SKetchyVQA \cite{sketchyvqa}, Visual Genome \cite{visualgenome}.
    \item OCR: DVQA \cite{dvqa}, ArxivQA \cite{arxivqa}, OCRVQA \cite{ocrvqa}, DocVQA \cite{docvqa}, ChartQA \cite{chartqa}, IconQA \cite{iconqa}, AI2D \cite{ai2d}, InfoVQA \cite{infovqa}.
\end{itemize}
\subsection{Video Instruction Tuning}
To enhance \model{}'s capability in handling videos, we further fine-tune \model{} on a mixture of our proposed datasets and the training splits of PerceptionTest and NextQA \cite{nextqa}. The video data is structured into a natural representation as follows: "1s: \texttt{<image>}; 2s: \texttt{<image>}; ..." This format incorporates explicit timestamps into the input, with \texttt{<image>} replaced by the compressed visual tokens before being fed into the LLM. We set the maximum context window to 16K during training. The number of tokens per frame is randomly sampled from the interval $ \left[ 16, \min\left( \frac{N_{\text{max}}}{T}, 576 \right) \right]$, where $N_{\text{max}}$ represents the maximum number of visual tokens—set to 12K by default for \model{}-7B and 10K for \model{}-14B. And the $T$ denotes the frame count of a video, which can be up to 256 in our datasets.

%% file: sec/4_experiments.tex
\section{Experiments}
\subsection{Deployment Details}
Unless otherwise specified, we apply the greedy decoding strategy for the language model. During inference, when the maximum number of frames is set to 
$T$
, we maintain a default sampling strategy of 1 frame per second if the video duration is less than 
$T$
 seconds; otherwise, we uniformly sample 
$T$
 frames. During inference, the count of tokens per frames is fixed as $\max\left( \frac{N_{\text{max}}}{T}, 576 \right)$.
\subsection{Benchmarks and Metrics}
We assess \model{} using a comprehensive range of established benchmarks for videos. 
In detail, we category the benchmarks we evaluate on into three classes:
\\
\textbf{1) Open-ended VideoQA.} This task is designed to prompt VideoLLM to provide free-form answers to questions related to video content. Our \model{} is evaluated on commonly used datasets: MSRVTT-QA \cite{xu2017msrvttqa}, MSVD-QA \cite{chen2011msvd}, Activity-QA \cite{yu2019activityqa}, and TGIF-QA \cite{jang2017tgif}. Following established protocols \cite{videochatgpt, videollava, videollama}, we utilize the GPT API for evaluation, assigning scores from 0 to 5. However, FreeVA \cite{freeva} observed that the comparison methods use different versions of the GPT APIs, which could significantly impact the final metrics. Therefore, we also specify the version of the GPT APIs used for comparison. It is crucial to mention that \texttt{GPT-3.5-turbo-0301} became unavailable in July 2024, thus, we report the performance of \model{} evaluated by \texttt{GPT-3.5-turbo-0613}. 
\\
\textbf{2) Multi-choice VideoQA.} In the context of a question and multiple candidate options, VideoLLM is tasked with selecting the correct option. Therefore, the performance of methods is evaluated based on Accuracy, defined as the ratio of selecting the correct answer. Compared with open-ended VideoQA, it requires the model has ability to follow the instruction to give the answer. Furthermore, the performance is not relied on external critical model, thus more stable and reproducible. We report the performance of \model{}  on VideoMME \cite{videomme}, MLVU \cite{mlvu}, TempCompass \cite{tempcompass}, EgoSchema \cite{egoschema}, and validation split of PerceptionTest \cite{perceptiontest}.
\\
\textbf{3) Multi-choice multi-image QA. } Similar to Multi-choice VideoQA, this task involves presenting a question alongside multiple candidates and several images. While the images may lack temporal relationships, they display more abstract correlations, requiring the model to extract information independently from each image. Although \model{} has not been trained on multi-image data at any stage, we evaluate its zero-shot performance on this novel task, which falls entirely outside the domain of \model{}. Not only are the types of images not seen during training, but the nature of the task itself is also unfamiliar. We assess performance on two public benchmarks: MuirBench \cite{muirbench} and MMIU \cite{mmiu}.
\subsection{State-of-the-art Comparisons}
\textbf{Evaluation on open-ended Video QA. }Following established protocols, we evaluate \model{} on various free-format video QA datasets to highlight its strong capabilities. As shown in Table \ref{tab:openvideoqa}, \model{} achieves state-of-the-art performance across the evaluated datasets. Notably, on the challenging ActivityNet-QA, \model{} significantly outperforms previous methods.\\
\textbf{Evaluation on multi-choice Video QA. }As shown in Table \ref{tab:openvideoqa}, our \model{} achieves SoTA performances on a wide range of benchmarks. \model{}-14B can even achieves better performance than those models with a much larger LLM.\\
\textbf{Zero-shot task transfer on multi-image QA. }Surprisingly, we find our \model{} achieves great success when dealing with new scenarios that are never observed in all training stages. \\
\subsection{Ablation Studies}
\begin{table}[ht]
\centering
\begin{tabular}{c|c|c}
\toprule
\multirow{2}{*}{\textbf{Compressors}} & \textbf{VideoMME} & \textbf{MSVD-QA} \\
\cline{2-3}
& w.o. subtitles & Acc./Score \\
\midrule
Token Merging & 51.6\% & 61.9/3.6 \\
Token Pruning & 47.6\% & 59.5/3.5 \\
Pooing & 52.0\% & 62.0/3.6 \\
\bottomrule
\end{tabular}
\caption{Ablation studies on the architecture of dynamic compressors show that among the three architectures examined, Adaptive Pooling delivers the best performance.}
\label{dynamic}
\end{table}
\begin{table}[ht]
\centering
\setlength{\tabcolsep}{2pt}
\begin{tabular}{c|c|c|c}
\toprule
\textbf{\# Tokens/F} & \textbf{\# Max Tokens} &  \textbf{\# Max Frames} & \textbf{VideoMME} \\ 
\midrule
36 & 12,000 & 333 & 58.7\%\\
64 & 12,000 & 187 & 59.4\%\\
100 & 12,000 & 120 & 60.9\%\\
144 & 12,000 & 83 & 59.7\%\\
256 & 12,000 & 46 & 59.3\%\\
\midrule
36 & 8000 & 222 & 59.3\%\\
64 & 8000 & 125 & 60.3\%\\
100 & 8000 & 80 & 59.6\%\\
140 & 8000 & 57 & 58.6\%\\
256 & 8000 & 31 & 58.6\%\\
\midrule
36 & 4000 & 111 & 57.9\%\\
64 & 4000 & 62 & 58.3\%\\
100 & 4000 & 40 & 58.2\%\\
144 & 4000 & 27 & 57.0\%\\
256 & 4000 & 15 & 56.3\%\\
\bottomrule
\end{tabular}
\caption{Ablation studies on token count per frame indicate that \model{} achieves the best performance when the token count per frame is set to around 64 or 100, regardless of the maximum number of visual tokens $N_{\textit{max}}$.}
\label{tokencount}
\end{table}
\noindent \textbf{Ablation studies on architecture of visual token compressor.} We explored how the architecture influences the performance of \model{}. As discussed earlier, there are three types of compressors that can represent one image with a flexible number of tokens. We implemented a simpler training process to assess the effectiveness of each architecture. All models were pretrained on llava-558k and then fine-tuned on a small subset of the proposed dataset (approximately 11K mixed data). We used VideoMME and MSVD-QA to evaluate the performance of the architectures. As shown in Table \ref{dynamic}, the results indicate that token-pruning methods fall short in scenarios requiring high-level understanding, while the simple adaptive pooling operation achieves slightly better performance than the token merging method. Therefore, we adopted pooling as default.\\
\textbf{Ablation studies on token count per frame.} As shown in Table \ref{tokencount}, when we fix the maximum number of visual tokens and increase the token count per frame, the performance on VideoMME (w.o. subtitles) initially rises before eventually declining. This trend suggests that there is an optimal balance between the number of frames and the tokens allocated per frame. Furthermore, when we maintain a constant token count per frame while increasing the maximum number of visual tokens, we observe a slight upward trend in performance. However, once the number of sampled frames reaches a certain threshold, simply adding more frames does not lead to improved performance, as shown in line 2, 7, 12 in Table \ref{tokencount}. This implies that merely increasing the number of frames, without paying attention to the detailed information within each frame, is not an effective strategy. To this end, we set the max frames as 120 in our other experiments. 

%% file: sec/5_conclusions.tex
%
\section{Conclusions}
In this paper, we present a high-quality synthetic video-text dataset and introduce a dynamic visual token compressor that adapts to varying video lengths. This dataset addresses the shortage of high-quality video instruction tuning data and supports the open research community. Our extensive experiments demonstrate that \model{} performs exceptionally well not only in Open-ended VideoQA and Multiple-choice VideoQA but also in Multi-image QA tasks. These contributions facilitate more effective processing of video data, bridging the gap between open-source and industry-level models while establishing new benchmarks for future research in this domain. Furthermore, our findings indicate that training exclusively on video data can enhance the performance of LVLMs on multi-image tasks, potentially inspiring future research to develop more robust training strategies for creating generalizable LVLMs across various tasks.